\documentclass{sig-alternate}

\usepackage{amsmath}
\usepackage{paralist}
\usepackage{booktabs}
\begin{document}

%

\title{Measuring Human-perceived Similarity in Heterogeneous Collections}

%
%
%
%
%

\numberofauthors{4} 
\author{
\alignauthor
Jesse Anderton\\
       \affaddr{Northeastern University}\\
       \email{jesse@ccs.neu.edu}
\alignauthor
Pavel Metrikov\\
       \affaddr{Northeastern University}\\
       \email{metpavel@ccs.neu.edu}
\alignauthor Virgil Pavlu\\
       \affaddr{Northeastern University}\\
       \email{vip@ccs.neu.edu}
\and  
\alignauthor Javed Aslam\\
       \affaddr{Northeastern University}\\
       \email{jaa@ccs.neu.edu}
}

\maketitle

\begin{abstract}
We present a technique for estimating the similarity between objects such as movies or foods whose proper representation depends on human perception. Our technique combines a modest number of human similarity assessments to infer a pairwise similarity function between the objects. This similarity function captures some human notion of similarity which may be difficult or impossible to automatically extract, such as which movie from a collection would be a better substitute when the desired one is unavailable. In contrast to prior techniques, our method does not assume that all similarity questions on the collection can be answered or that all users perceive similarity in the same way. When combined with a user model, we find how each assessor's tastes vary, affecting their perception of similarity.
\end{abstract}


\section{Introduction}
Many common tasks in Data Mining, Machine Learning, Information Retrieval and related fields depend on an embedding
of data objects into some feature space, either explicitly through numeric features or 
implicitly through a pairwise similarity function (kernel). A significant amount of work has 
gone into developing the correct feature space for a given task, often attempting to 
model user intentions, or model a notion of similarity between objects in the collection.
 For instance, density based methods or SVMs depend explicitly on similarity (or kernel) mapping; in collaborative filtering, a product/item recommendation often depends on how similar a product is to other products the user has purchased, or products that other users similar to the potential customer have purchased.
	Although similarity is central to many of these tasks, and great strides have been 
made in producing automatic measures of similarity for many tasks~\cite{4072747,Shi:2010:MMM:1869652.1869658,Guo:2011:IQS:2063576.2063619}, we find that 
the ability to directly measure a human perception of the similarity between objects in 
a collection is currently lacking. This is unfortunate because the success of many 
learning tasks with realistically difficult data depends on the ability to accurately model user 
satisfaction or preference, particularly in cases where the users' notion of similarity 
inherently depends on subjective information.


We propose a framework allowing researchers to construct pairwise 
object similarity functions for domains that require subjective or user-specific 
information to achieve high accuracy. These similarity functions, or similarity 
kernels, can be used to
(1) run \emph{learning algorithms} on raw data such as movies; 
(2) \emph{analyze} an existing set of features, pointing out deficiencies in matching human judgements and providing guidance toward developing more natural features;
(3) model \emph{user differences in perception}; or
(4) \emph{evaluate} a variety of tasks such as Information Retrieval rankings, 
query diversity, and product recommendation.
Rather than imposing a set of features or a similarity function explicitly upon a data set, we propose to instead learn either a pairwise object similarity function or, equivalently, an embedding of those
objects into a metric space which best fits the observed human-generated similarity information.
Different collections of objects, users, and tasks pose different challenges:
\begin{compactitem}
\item to define what is meant by \emph{similarity} in a given task or context,
\item to collect \emph{human perception} as feedback from appropriate questions / answers,
\item to design \emph{backend likelihood models} for human response actuation on the learning model for similarity, and
\item to use \emph{optimization/training procedures} to infer optimal model parameters.
\end{compactitem}
Existing methods  currently fall short in one or more of these areas, preventing their use in 
more complex data collections. The first difficulty is that humans are very inconsistent in assessing similarity, so quantitative questions like ``\texttt{how similar is X to Y?}'' usually are not good user feedback~\cite{carterette2008here}. For most tasks, it has been shown that preference questions like ``\texttt{is X more like Y, or more like Z?}'' make for better, consistent, feedback~\cite{Kalai11}; but even then, similarity is not at all straightforward. For example, if we know a user rates the dish \texttt{eggplant parmesan} very highly, 
which of the following is more similar to it: \texttt{baingan bharta} (an Indian eggplant dish), 
or \texttt{lasagna} (an Italian dish)? While the first is closer in terms of ingredients but quite different meal
experience, the second is closer in terms of cuisine but different in ingredients and nutrition.
Thus a correct similarity notion has to account for who is making the preference, and possibly
for the contexts in which these food dishes are discussed.
A movie example: If Joe wants to see the movie \texttt{Star Wars}, which is unfortunately unavailable, would he prefer \texttt{Lord of the Rings} or \texttt{The Matrix} as an alternative? Not only is it hard to answer such a question solely from superficial features like actors, genre, director etc; different users will legitimately answer differently, so even with good movie features, the models have to account for user personality or taste on movies. In short, high accuracy depends on measuring subjective perception.

\subsection {Omer Tamuz et. al 2011 Kernel paper}
Omer Tamuz et al presented in their 2011 ICML paper ``Adaptively Learning the Crowd Kernel''\cite{Kalai11} an \emph{original framework} for doing just that: learning similarity from human input in the form of preferences ``is X more like Y or more like Z''?. This allows characteristics of an image and
relationships between the images to be inferred from feedback, and
not imposed artificially. They demonstrated good results on pictures of ties and of human faces. Our study is a follow-up on their excellent idea. 

After discussion with the authors and reimplementation of the procedures, we found several shortcomings: (1) the image data is not the most telling for assessing performance: the applicability of such framework is rather useful on data objects for which human perception is subjective and far less computable than it is on images (an automatic method for face detection or tie pattern recognition can do quite well); (2) their crowd answers are limiting the human input by not allowing certain natural answers ``I dont know,'' ``neither,'' ``both,'' etc. --- we found that  these answers are returned a significant amount of the time, and that not explicitly using them as feedback slows down the learning process; (3) the original framework does not allow for any contradiction or associative disagreement between answers, hoping that the kernel eventually averages out the feedback --- but subjective feedback on objects like movies or foods or even text does contain large disagreement, thus better handled explicitly by accounting for various users; (4) the optimization method given, based on gradient descent, is slow and quite inefficient at finding a global optimum for any number of data dimensions more than two.

Section~\ref{sec:method} describes the improved kernel learning framework as compared to the original 2011 paper\cite{Kalai11}, making it far more suitable for the kinds of data and problems targeted:
\begin{itemize}
\item complex datasets with subjective feedback, such as snippets' relevance, paintings, movies, songs, food dishes etc, are likely the datasets one would apply this framework to, as opposed to tie patterns or human faces used in the past to demonstrate the results;
\item humans' possible answers are expanded to include more natural answers for preference questions, and consequently the new learning framework takes such answers into account when fitting the model. Additionally a notion of data objects ``too far away to be comparable'' is modeled into the probabilistic framework;
\item multi-dimensional optimization is sped up by an order of magnitude over gradient descent methods, using a Quasi-Newton/Hessian numerical approach; and
\item user models are embedded in the framework. 
\end{itemize}
Section~\ref{sec:users} presents an adaptation of the kernel framework to embed user models. It greatly improves the kernel fit onto observations by allowing disagreement across users, something not possible in the original framework. It also makes better predictions given a user model that matches the test observations, for example when the same user is known to have contributed to the training observation set, or when the testing set user profile is known as a combination of trained users.   

\begin{figure}[h]
\centering
\includegraphics[width=0.48\textwidth]{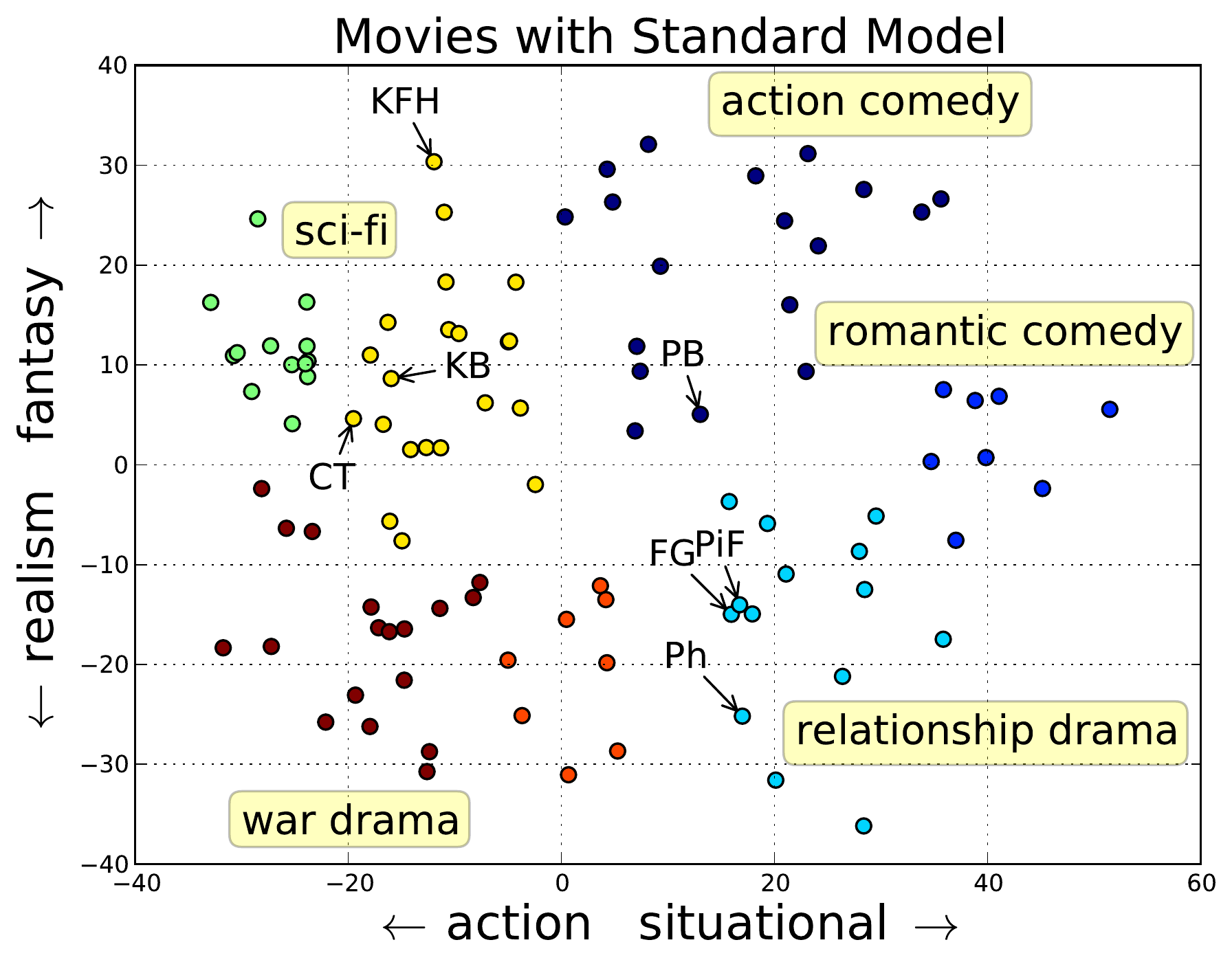}
\quad
\includegraphics[width=0.48\textwidth]{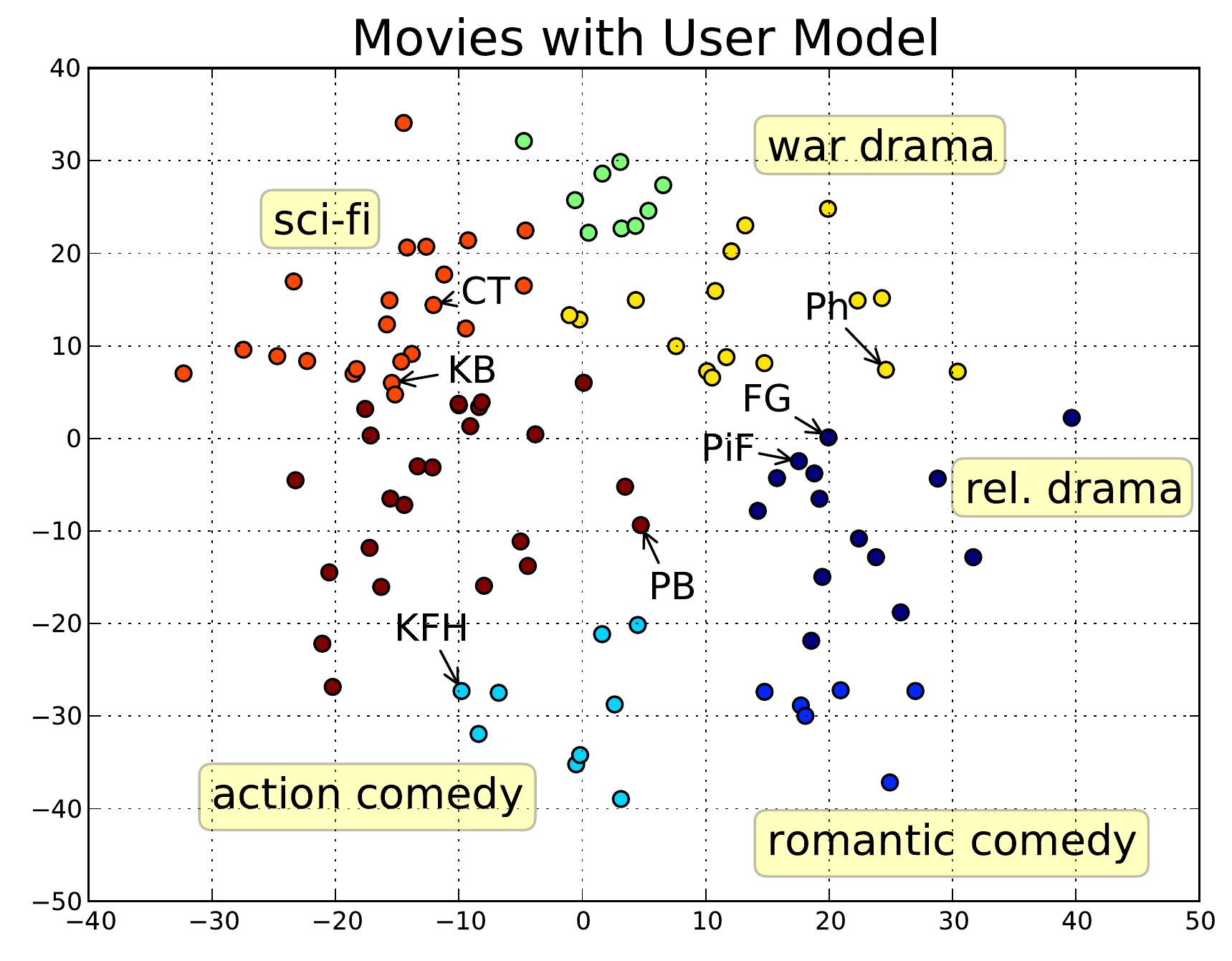}
\caption{Movies in 2D, without user model (original, top) and with user model (new framework, bottom). Axis and genre labels are approximate, meant to give a sense of movie positions within the space. Note that the user model axes are a linear combination of the standard model axes, but relative positions are consistent.}
\label{movies-2d}
\end{figure}

Figure~\ref{movies-2d} illustrates the results of the data collection process after several grad students tested the feedback interface with preference questions about movies, before running the Mechanical Turks study presented below. The actual object positions are revealing beyond mere cluster membership. For instance, ``\texttt{The Princess Bride}'' (\texttt{PB} in the figure) falls in the ``action comedy'' cluster along with spy/cop movies, but is positioned very near the ``romantic comedy'' cluster. ``\texttt{Forrest Gump}'' (\texttt{FG}) shares a cluster with both ``\texttt{Pay it Forward}'' (\texttt{PiF}) and ``\texttt{Philadelphia},'' (\texttt{Ph}) but is much nearer the upbeat former than the more dramatic latter film. ``\texttt{Crouching Tiger, Hidden Dragon}'' (\texttt{CT}) is very near ``\texttt{Kill Bill: Vol. 1}'' (\texttt{KB}) and more distant from the comedic ``\texttt{Kung Fu Hustle},'' (\texttt{KFH}) though all three martial arts films are fairly near each other.

\subsection{Other Related Work}


%

\textbf{Human Input and Image Similarity} 
Considering the space of image similarity, we see that a large amount
of work has been done to perform human relevance feedback for
content-based image
retrieval~\cite{chen2001one,cox1998optimized,rui1998relevance}. This
is mainly because a machine has a harder time deducing the content of
an image than of text, and as such, human judgments are far more
valuable than machine features.  Human assessments for images are far
easier than for text, however, because humans have a harder time
processing an entire document than a single image.  

There are, however, many parallels
that can be drawn between image and text feedback.  For both image and
text retrieval, the user model can be thought of in two ways: a user
can search for a category of information or
images~\cite{chen2001one}, or for a target piece of information or
image~\cite{cox2000bayesian}. In the former case, the feedback is
binary---whether or not the document or image fits in the category;
the latter is a continuous measure of how close the document or image is to the
target.  It is the latter user model which we use in the similarity
kernel, as relative judgments have been shown to have less bias and
therefore less noise than absolute
judgments~\cite{carterette2010effect,ipeirotis2010quality,voorhees2000variations}.
Additionally, relative judgments allow for arbitrarily fine-grained
notions of graded relevance~\cite{carterette2008evaluation}, or in our
case, notions of relevance across multiple dimensions.

\textbf{Preferences vs.\ Nominal Grades on IR document assessment.} 
Some work is being done using preference pair judgments for
retrieval~\cite{carterette2008here} or preference triples~\cite{chandar2012using}, but these preferences either
require many judgments or focus only on ordering documents for
relevance and diversity; they do not examine the similarity between
objects.  

The current approach to evaluating search engine diversification is subtopic relevance~\cite{Agrawal09,Clarke08}. However, these subtopics are created in advance by positing query facets or
manually inspecting query logs for query variants.  This is
artificially imposing the dimensions of relevance, which as such do
not necessarily correlate to the idea of relevance held by a group of
users~\cite{Golbus}.  Additionally, the documents for a query are
crawled and pooled with no knowledge of the subtopics, and as such do
not necessarily match them.


%
%

\section{Modeling answers for heterogeneous data}
\label{sec:method}

The model from Tamuz et al. asks similarity questions by selecting some object triple $(a, b, c)$
from a fixed collection of $n$ objects and asking, ``is object $b$ or $c$ more similar to $a$?''
It uses the responses to position each object within a $d$-dimensional metric space $M$,
or equivalently a $n \times n$ similarity kernel $K=MM^T$. The model assumes that there
is some natural feature space for these objects which users intuitively employ to compare them,
and that if you get the answers to enough similarity questions you can find not only the true
similarity value for any object pair but also the natural features human assessors use when
comparing them.

While we do not contest this assumption, we find that forcing users to choose between
$b$ and $c$ introduces unnecessary noise for more heterogeneous object collections.
When objects in these collections are relatively dissimilar, a human assessor is often
unable to estimate their similarity. Our initial experiments with text snippets on the original model suffered
from this problem; our assessors often complained that the questions didn't make
sense, as the three objects they were comparing seemed unrelated.
Worse, forcing a preference between the three objects
when $b$ and $c$ are both quite dissimilar from $a$ produces noisy observations:
the model will prefer to keep $a$ close to whichever
object was preferred, when in fact neither should be near $a$. If these noisy
observations outnumber the more meaningful ones, the model may never converge on a good solution.

We address this problem by providing users with a third option and updating our model
to use this extra information. We still ask the question, ``is $b$ or $c$ more similar
to $a$?'' However, in addition to expressing a preference we permit the users to say,
``Neither is similar to $a$.'' We interpret this answer
to mean that both $b$ and $c$ are somewhat distant from $a$ within the metric space $M$.
We make the modeling assumption that there exists some distance $d$ beyond which
users lose the ability to estimate object similarity, and that if both $b$ and
$c$ are at least this distance from $a$ then assessors are very likely to select ``neither.''
Note that we cannot infer from this answer that $b$ and $c$ are similar to \emph{each other}:
they may be on opposite sides of $a$, each dissimilar for its own reasons.
The value of $d$ could reasonably be fit on a per-object, per-user, or global basis. For our
experiments in this section, we fit the parameter globally. Later we show a variation that
fits the parameter per-user.

Using $\delta_{ab} =||M_a-M_b||^2=K_{aa} + K_{bb} -2K_{ab}$ and the similarly defined $\delta_{ac}$
to denote the squared Euclidean distance from
$a$ to $b$ and to $c$ within the embedding $M$, and with smoothing parameters $\mu$ and
$\lambda$, we
give the probability of the user choosing ``neither'' as shown in Table~\ref{tbl:prob}.

The probability of the user preferring either $b$ over $c$ or vice-versa as a substitute for $a$ is computed accordingly. Note in Table~\ref{tbl:prob} that the original 2-answer Kernel cannot, by design, produce a probability of the user choosing ``neither.''


\begin{table}[htb!]
\centering
$\begin{array}{c|l|l}
 \toprule
\mathbf{Answer} & 
\mathbf{Kernel} & 
\mathbf{Kernel} \\
\mathbf{prob.} & \mathbf{two\text{ }answers} & \mathbf{three\text{ }answers}\\
 \midrule
\hat{p}^a_{bc} & 
  \begin{aligned}
    \frac{\lambda + \delta_{ac}}{2\lambda + \delta_{ab} + \delta_{ac}}
  \end{aligned} &
  \begin{aligned}
	  \left( 1 - \hat{p}_{neither} \right)
	  \cdot
	  \frac{\lambda + \delta_{ac}}{2\lambda + \delta_{ab} + \delta_{ac}}
  \end{aligned} \\
& &\\
\hat{p}^a_{cb} & 
  \begin{aligned}
    \frac{\lambda + \delta_{ab}}{2\lambda + \delta_{ac} + \delta_{ab}}
  \end{aligned} &
  \begin{aligned}
	  \left( 1 - \hat{p}_{neither} \right)
	  \cdot
	  \frac{\lambda + \delta_{ab}}{2\lambda + \delta_{ac} + \delta_{ab}}
  \end{aligned} \\
& &\\
\hat{p}_{neither} &
  \begin{aligned}
	  0\text{ (N/A)}
  \end{aligned} &
  \begin{aligned}
	  \frac{\mu + \delta_{ab}}{\mu + d^2 + \delta_{ab} }
	  \cdot
	  \frac{\mu + \delta_{ac}}{\mu + d^2 + \delta_{ac}}
  \end{aligned} \\
 \bottomrule
\end{array}$
\caption{Probabilistic model for Kernel likelihood. $\hat{p}^{a}_{bc}$ here means the estimated probability of the user claiming $b$ is more similar than $c$ to $a$. The original two-answer Kernel is a particular case of the three-answer Kernel with $\hat{p}_{neither}=0$ (equivalently $d=\infty$).}
\label{tbl:prob}
\end{table}

Given this model
and some positioning of the objects within the kernel information space, we can calculate
the probability of a user responding to a similarity question in any given way. This allows
us to find an optimal information space which maximizes the probability of the actual
user responses. For our experiments we set $\lambda = 1$ and allow $\mu$, $d$, and the
object positions to be chosen to maximize the probability of the assessors' observations.

To be specific, given a set of $n$ observations $\{(a_i, b_i, c_i, r_i)\}$ over objects $a_i$, $b_i$, and $c_i$ and response $r_i$,
we seek to minimize the expected log loss as a function of all object embeddings $\{M_x\}$ and a set of parameters $\Omega = \{\mu, d\}$
\[ L\left(\{M_x\}, \Omega \right) = - \sum_{i = 1}^{n} \lg \hat{p}(r_i | M_{a_i}, M_{b_i}, M_{c_i}, \Omega). \]

The optimization surface is non-convex and quite ``bumpy,'' and finding a global optimum is challenging.
Fortunately, we find that the results are good even for the local optima we converge on. We have employed
a quasi-Newtonian/Hessian method which performs gradient descent on the loss function using learning
rates computed from the second derivative of $L$, with multiple random initialization points to find
a reasonably good local optimum.

\subsection{Experiments and Results}

To test our method, we crowdsourced assessments for two data sets, described below. We compare
our three-answer model to the two-answer model described by Tamuz et al. We ran 20 rounds of active learning
independently for each model and data set. In each round, we asked crowd workers one question with each
object in the collection serving as $a$ in an ($a$, $b$, $c$) triple.
We sent the questions to Amazon Mechanical Turk (AMT), asking 12 questions per HIT.
These questions were divided between active learning questions selected by the model being trained,
randomly sampled test questions used for evaluation, and trap questions used for accepting or rejecting
a batch of crowdsourced answers. The trap questions were selected using a manual clustering of the objects,
with either $b$ or $c$ drawn from the same cluster as $a$ and the other object drawn from a different
cluster. A HIT was accepted if the worker selected the object from the same cluster as $a$ for at least
one of the two trap questions. This serves as a mild quality filter, since even totally random workers
can be expected to pass the trap questions fairly often. In practice, we observed about a 90\% acceptance
rate of submitted work.

Each HIT for the baseline model included 10 active learning questions and 2 trap questions.
For our three-answer model, we asked 2 trap questions, 5 active learning questions, and 5 test questions per HIT.
We mixed training and test questions in the same HIT so we can later evaluate the predictive power of
our user model, described in the next section. Further, the test questions were only asked for
the three-answer model because we interpret an answer of ``neither'' on a test question as meaning that the
comparison is uninteresting, or at least unreliable; we excluded these test triples from our evaluation.

\paragraph{Movie data}
Our first data set is a collection of 100 popular movies. We decided to measure
the extent to which a user who wants to see one movie would be satisfied with some other movie.
For instance, perhaps they have a movie in mind but have to select from
a limited menu which doesn't include their movie of interest.

\begin{figure}
\centering
\includegraphics[width=0.5\textwidth]{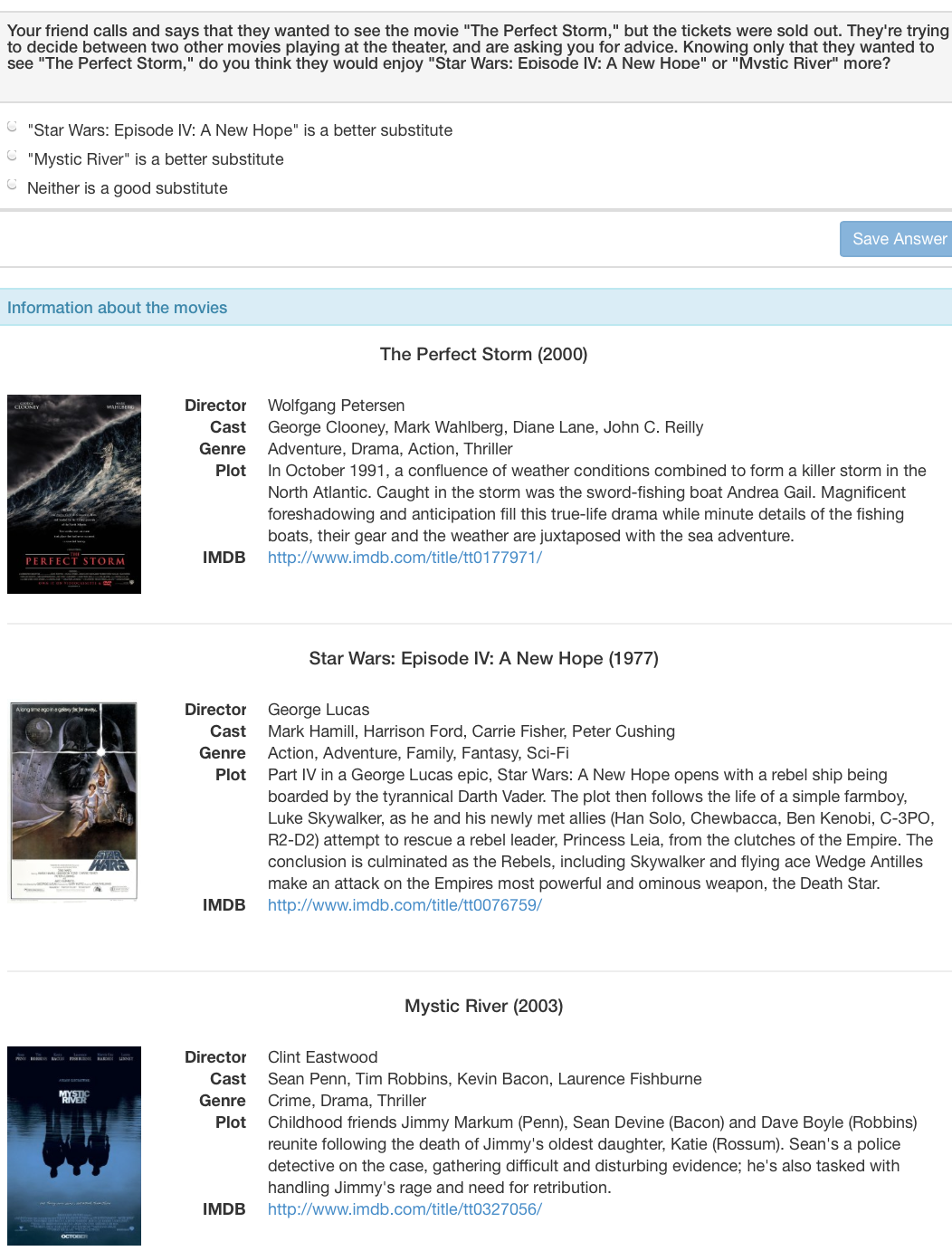}
\caption{Preference Feedback interface shown for movies.}
\label{fig:gui}
\end{figure}

We chose movies because many people are familiar with the task of choosing a movie,
but also because there is enough variety among them that many movies are sufficiently dissimilar
that they seem to have nothing to do with each other. For instance, if you want to see a
light martial arts comedy you are presumably very unlikely to choose an intense relationship drama
as a replacement. This collection
can thus be expected to have the property that forcing a user to choose a replacement
for any given triple $(a,b,c)$ will introduce noise into the data: if given the option, for at least
some of these triples, the user would simply never
replace $a$ with either $b$ or $c$. Comparing movies also involves a large subjective element.
For instance, some viewers seem to put the most weight a movie's genre and special effects.
Other viewers pay less attention to those features, and think about the type of story that's
being told and the movie's overall theme or message. We expect that this can lead to natural disagreements,
where people choose different movies as suitable replacements based on their personal points of view.

The movies in our collection were selected to fit into a few large scale genres (e.g.
comedy, drama, or science fiction), each with several sub-genres (such as war drama,
relationship drama, or a personal quest) and some genre-crossing features
(relationship drama vs. romantic comedy, or serious action versus comedic action). We 
constructed a web-based interface which presents the movies' titles, posters, 
director, genres, main actors, and summaries. Users are shown three movies at a 
time along with the following instructions:

\emph{Your friend calls and says that they wanted to see the movie ``Movie A,'' 
but the tickets were sold out.  They're trying to decide between two other movies 
playing at the theater, and are asking you for advice.  Knowing only that they 
wanted to see ``Movie A,'' do you think they would enjoy ``Movie B'' or 
``Movie C'' more? Note that this is not a question about which movie you 
like more, but instead a question about which movie is most similar to ``Movie A.''}

\paragraph{Food data}
Our second data set repeats the same experiment using a different collection of objects.
In this case, we used 100 different types of food. We chose a similar similarity question
to produce comparable results: if someone wants food $a$ but must instead choose between foods
$b$ and $c$, which would be a better replacement? We expected foods to be easy for crowd
workers to compare, but also to have triples where neither option is similar at all to $a$
and to have subjective answers. For instance, a vegetarian can be expected to have different
views from a non-vegetarian when trying to replace a vegetable-based entree.
Our interface presented the food name, a photo, and a description drawn from the food type's Wikipedia article -- usually 
its first paragraph. We selected mainly dinner entrees from various cultures, 
with many themes linking different kinds of foods: main ingredient, ethnicity, 
style (soups, dumplings, sandwiches, etc.) and so on. As with movies,
we ran a separate active learning process for each of the two models we compare.

This data set proved more challenging than the movies data.
The most important contributing factor may be that foods don't fall as neatly into pre-defined clusters.
When comparing movies, users seem inclined to compare based on genre first, and then based on
similarity within genre. Foods don't have such clear-cut 
categories. There are many ambiguous foods which sit somewhere between the natural groups.  (For instance,
is a hamburger more like a ham sandwich or like a steak?) We believe this led to less consistent answers, both
between assessors and for observations from the same assessor.

\subsection{Results for Updated Model}

\begin{figure}[t!]
\centering
\includegraphics[width=0.5\textwidth]{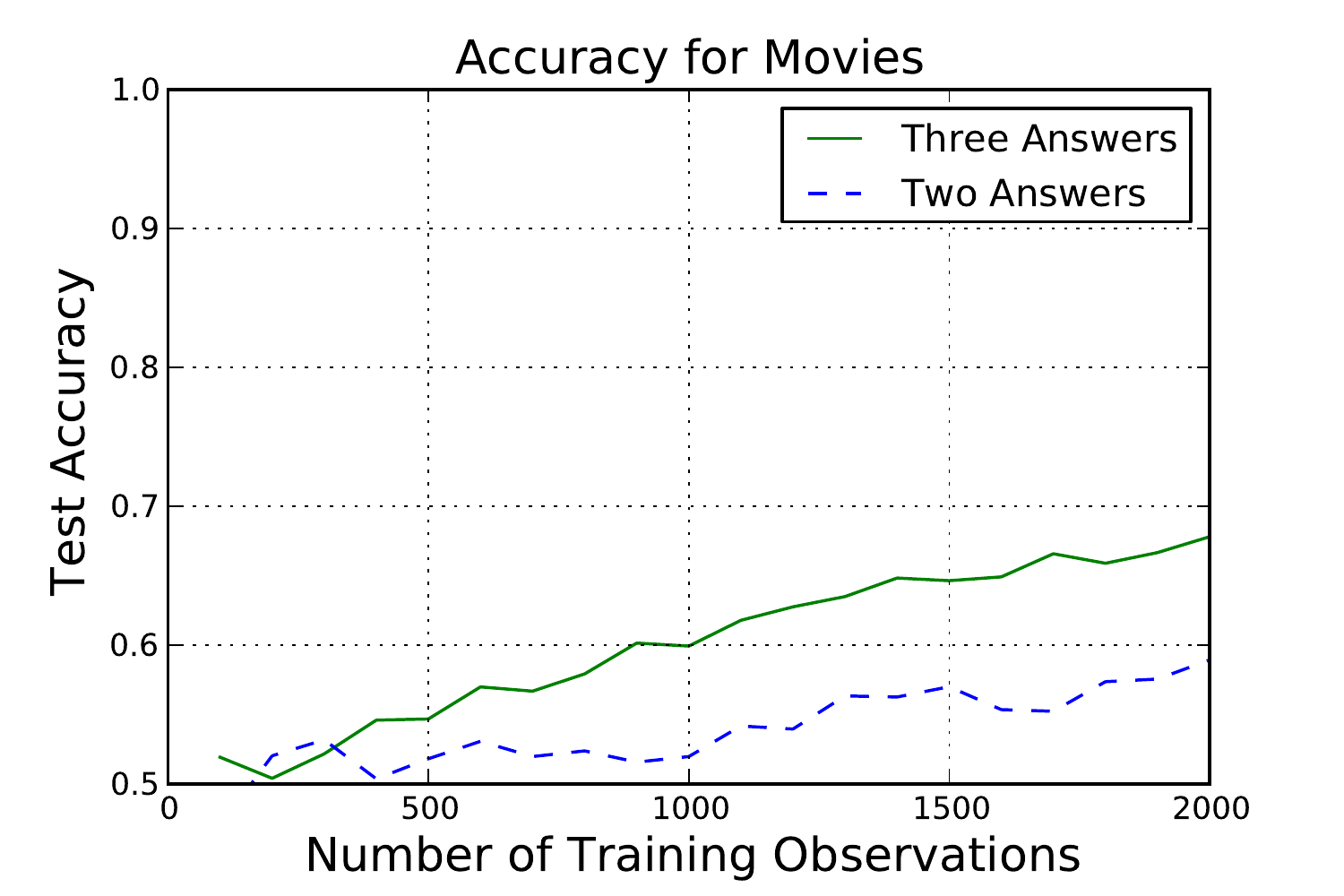}
\quad
\includegraphics[width=0.5\textwidth]{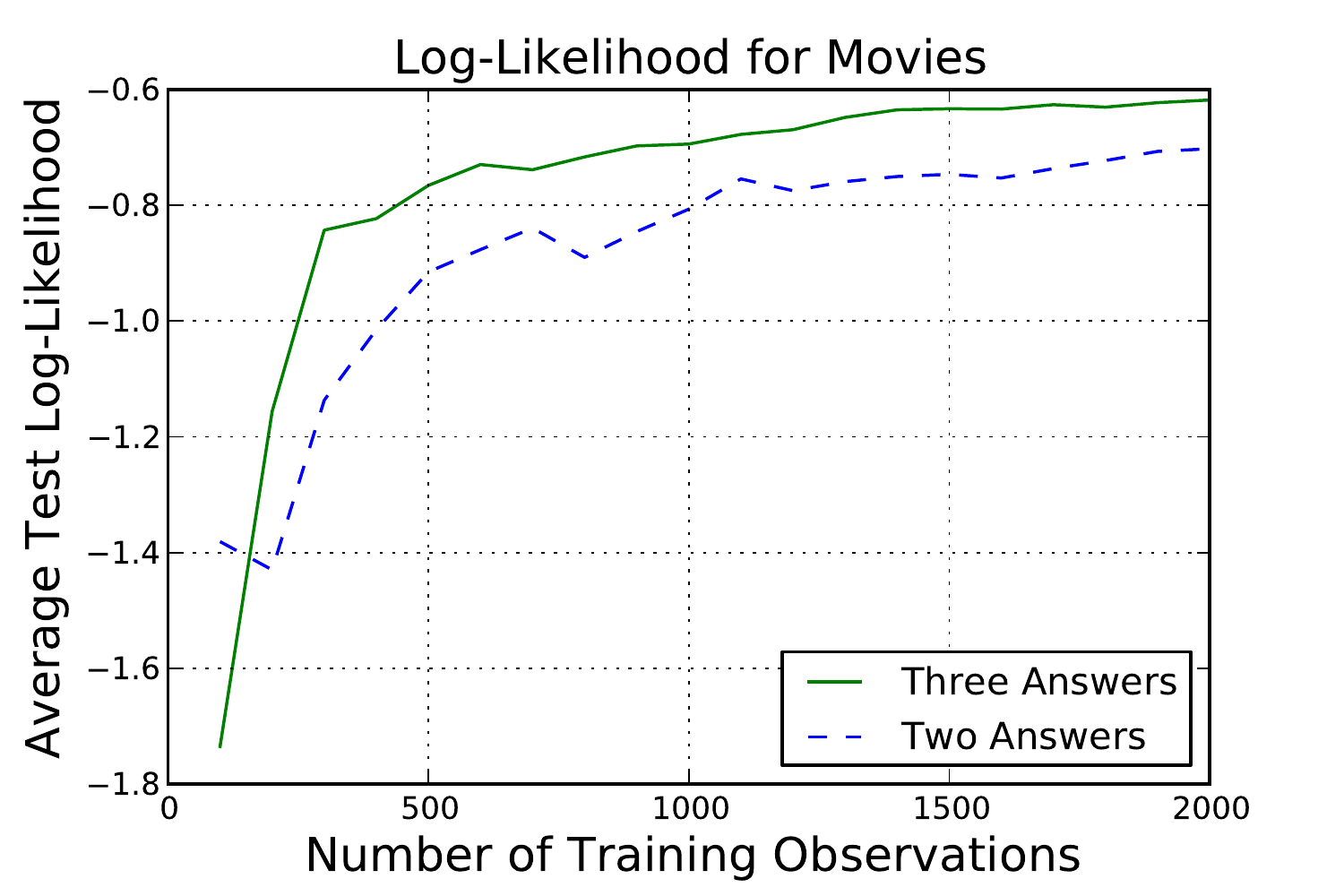}
\caption{Learning curves for baseline ``two-answer'' kernel and our ``three-answer'' kernel on movie data.
An accuracy of $0.5$ corresponds to random guessing, and due to inconsistency of opinions we don't believe
an accuracy of $1$ is achievable. Log loss is computed from the two-answers model using the kernel output
from each active learning process, so that probabilities are comparable. Although the log-loss reveals that
both models are learning, the baseline struggles to improve much beyond random accuracy. Our model shows
a clear improvement, learning at an approximately linear rate.}
\label{baseline-accuracy-movies}
\end{figure}

\begin{figure}[t!]
\centering
\includegraphics[width=0.5\textwidth]{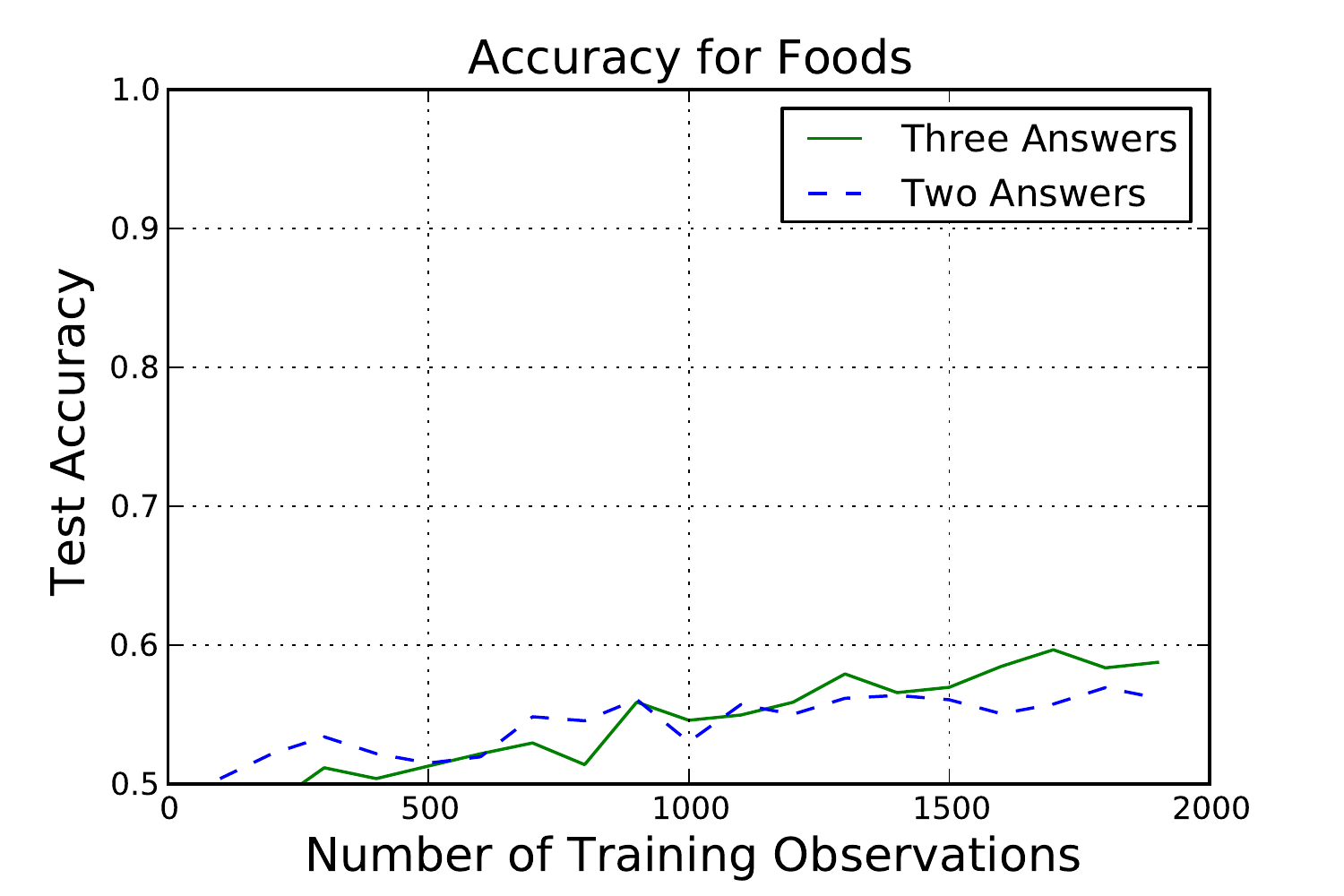}
\quad
\includegraphics[width=0.5\textwidth]{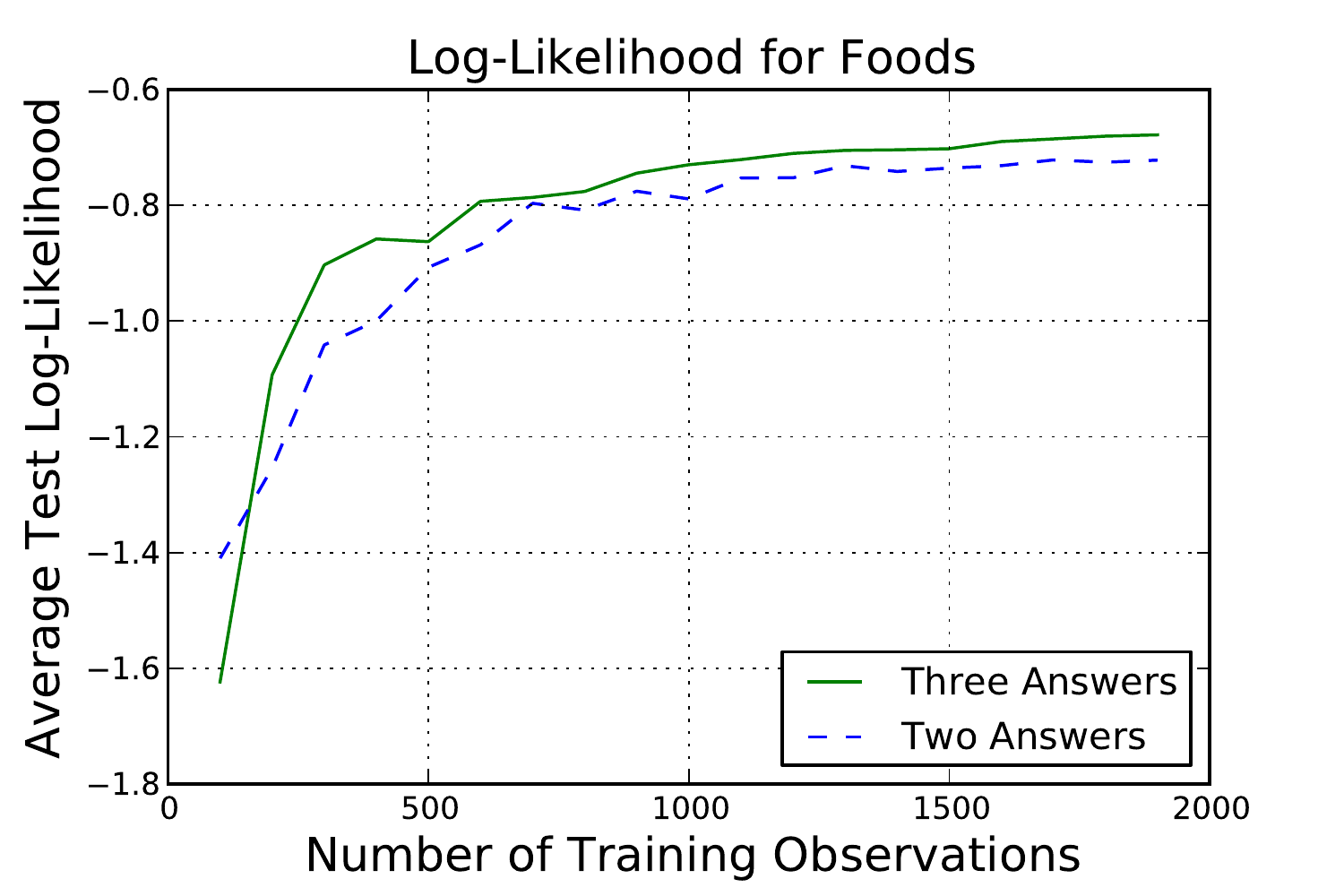}
\caption{Learning curves for baseline ``two-answer'' kernel and our ``three-answer'' kernel on foods data.
Log-likelihood of test data is consistently better for our model, but does not provide a clear benefit in
terms of accuracy.}
\label{baseline-accuracy-foods}
\end{figure}

\paragraph{Evaluation}
We compare the kernels output from each of the two active learning processes by evaluating them
against the test data. For any given test observation in which the user expressed a preference,
stating that either $b$ or $c$ is more similar to $a$ than the other, we consider the kernel to
be correct when the preferred object is more similar to $a$ and incorrect otherwise. This allows
us to build the learning curve in Figure~\ref{baseline-accuracy-movies} and Figure~\ref{baseline-accuracy-foods},
showing accuracy as a function of number of observations collected.

%
%

\section {User-specific Similarity}
\label{sec:users}

In the prior section, we showed how to relax the baseline model's assumption that all
objects in the collection can be compared by the assessors. Here we relax a further
assumption: that all users answer similarity questions consistently with each other.

The proper representation for objects in many learning tasks has a subjective component.
That is, users will disagree with \emph{each other} when answering similarity questions, but
will answer consistently with \emph{themselves}. This can pose interesting challenges in a
machine learning setting, where assessor disagreement can look like label or feature noise.
In this work, we approach the problem by first assuming that there is an objective representation
of the objects that all users can observe. This representation can be expressed as a \emph{global kernel}:
an embedding $M$ of the objects into a $d$-dimensional metric space shared by all users,
exactly as in the prior models. However, each user has his or her own scaling of the dimensions
of this metric space. This scaling, which we represent as a diagonal matrix $U^k$ for each user $k$,
produces a subjective \emph{personalized kernel} $K^k = MU^kM^T$ for each user. The global kernel,
used in our prior three-answers model, is equivalent to a personalized kernel with an identity matrix
for $U^k$.

To draw an example from the movies data set, consider that movie watchers enjoy different movies
for a wide variety of reasons. They see the same movies, and likely observe more or less the same
features of those movies, but with different sensitivities to those features. One person will pay
particular attention to the mood of the film, while another is interested in the special effects.
When comparing different movies, these viewers will each emphasize the features they care more about,
applying a larger scale to those features as compared to the features which are less important to them.

We update our three-answer model to draw distances from personalized kernels instead of the previously-used
global kernel. Where the three-answer model uses distances $\delta_{a,b} = \|M_{a} \cdot M_{b}\|_2^2$,
we instead use $\delta_{a,b}^k = \|M_{a} \cdot diag(U^k) \cdot M_{b}\|_2^2$. We also train
the distance $d_{neither}$ per-user instead of training a single global value to account for different
users' abilities to perceive similarity between objects.

This approach has a few potential problems. We would like the global kernel, parameterized with $U^k$ set
to an identity matrix, to be a reasonable prior for some user about whom we have no information. However,
the model as stated has no such constraint.
To see this problem, observe that the distance between two objects in a particular user's personalized
kernel is a function both of the global object positions in $M$ and the user's scaling in $U^k$. If every
value along some dimension $d$ of $M$ were reduced by half and the value of $U^k_{d,d}$ was doubled for
every user, the distances between objects would not be affected.
This has no impact on the predictive power of personalized kernels, but it means that the information
in the global kernel is somewhat deficient.
Second, we have a varying amount of information per user, ranging from five to hundreds of assessments
from particular users in our training set. This is relatively common in practice, both in a commercial setting
(where some customers rate only a few items, while others rate many) and in crowdsourced data
(where a worker may or may not choose to accept additional tasks). Our ability to accurately fit a personalized
kernel is dependent on the number of assessments available, and with few assessments it's easy to overfit the user parameters.

We resolve these problems by imposing a prior on our user kernels. We want this prior to have a mode of 1,
to keep the identity user for the global kernel meaningful, and to assign zero probability mass to negative values,
so mirroring dimensions relative to other users isn't possible. We accomplish this by putting a Gaussian
prior with mean zero and per-dimension standard deviation $\sigma_d$ on the logarithm of the values of
each $U^k_{d,d}$. This makes our loss function

\begin{align*}
L\left(\{M_x\}, \{ \Omega^k \} \right) = &-\sum_{k,d} \mathcal{N}(\log U^k_{d,d}, \sigma_d) \\
 &- \sum_{i = 1}^{n} \lg \hat{p}(r_i | M_{a_i}, M_{b_i}, M_{c_i}, \Omega)
\end{align*}

with $\Omega^k = \{ \mu^k, d^k \}$ parameterized per user.

Although you could imagine learning $\sigma_d$ per dimension,
justified by the intuition that variance of user feature sensitivity is likely to differ from feature to feature, for
our results here we simply fix $\sigma_d = 0.18$ for all dimensions.

\subsection{Results for User Model}

We evaluate our user model on the same data described in the prior section. We compare three
different methods of obtaining similarity predictions:
\begin{itemize}
\item the \emph{three-answer kernel} used in the prior section,
\item the \emph{general kernel} trained with user-specific parameters but making predictions
for an identity user ($U^k = I$), and
\item the \emph{personalized kernels} trained with user-specific parameters, and making predictions
for test data observations using users' own personalized kernels.
\end{itemize}

While the three-answer kernel outperforms the original two-answer kernel, 
both the general and personalized kernels outperform the three-answer kernel. This helps
validate our hypothesis that variation between users can be effectively modeled as a scaling of features.

\begin{figure}[t!]
\centering
\includegraphics[width=0.5\textwidth]{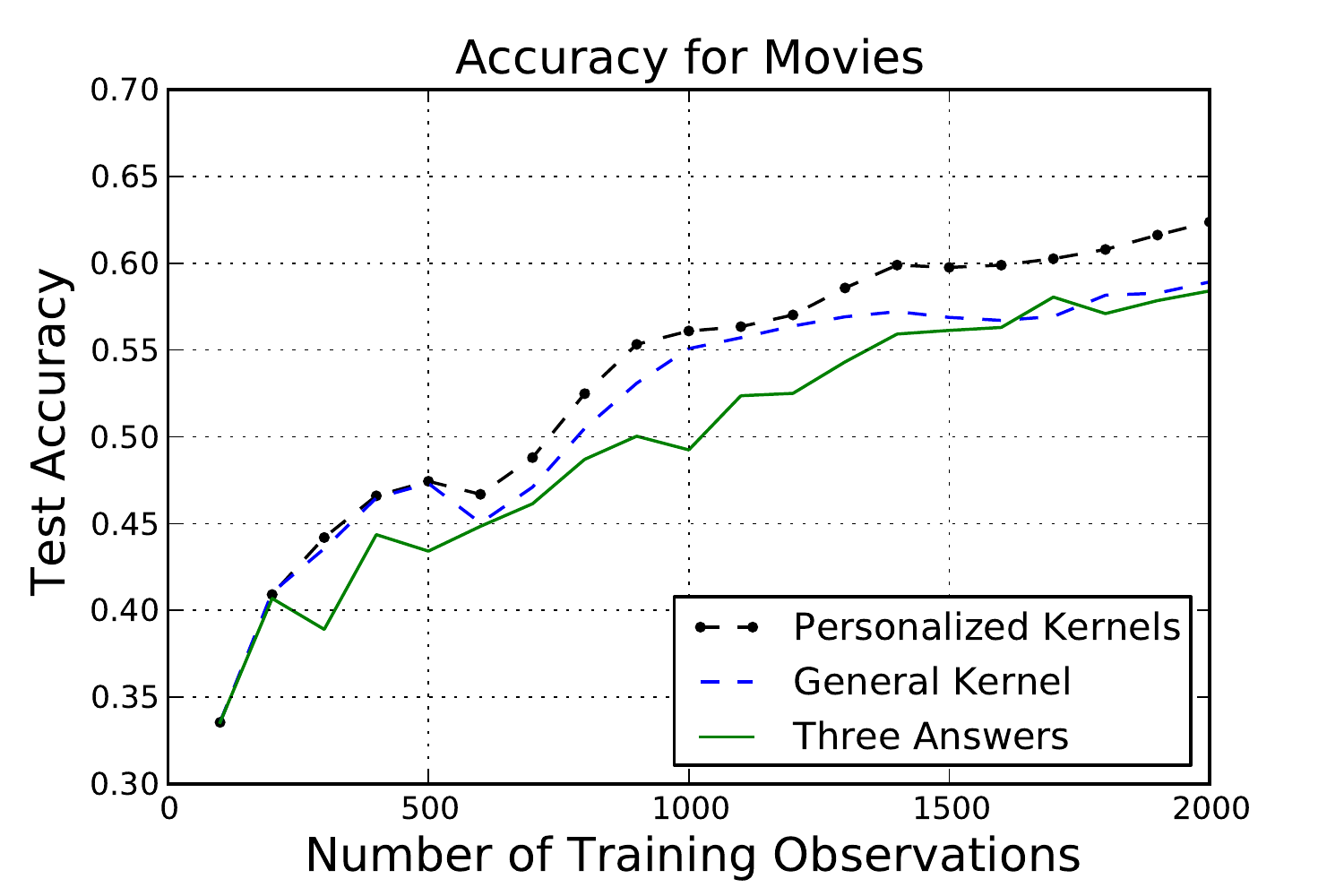}
\quad
\includegraphics[width=0.5\textwidth]{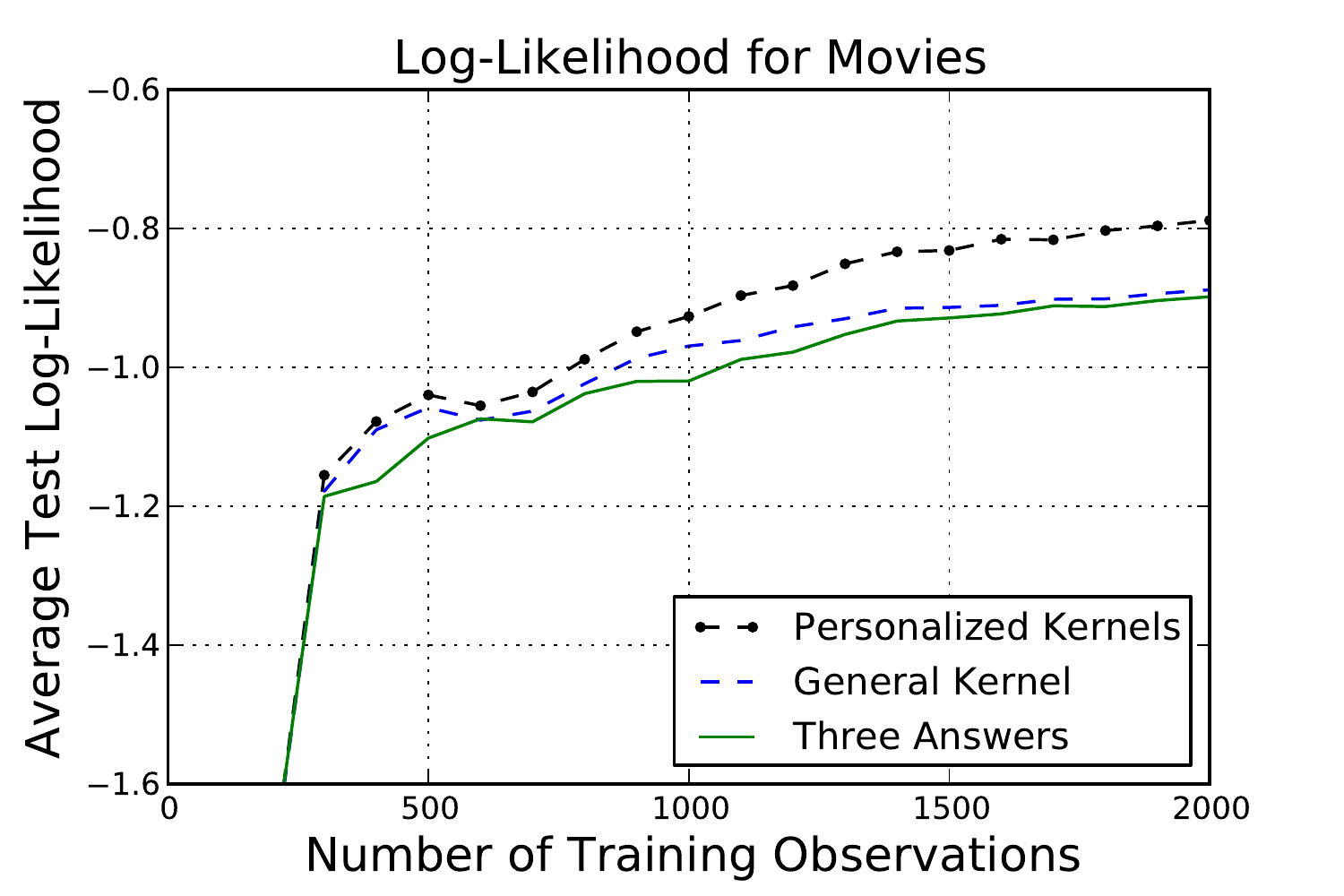}
\caption{Learning curves for user models on movies data.
Note that these plots are not directly comparable to the two answer baseline, because all the models
here were allowed to predict ``neither.'' A random answer lies somewhere around 1/3 probability, depending on the
prior value of $d$.}
\label{user-accuracy-movies}
\end{figure}

\begin{figure}[t!]
\centering
\includegraphics[width=0.5\textwidth]{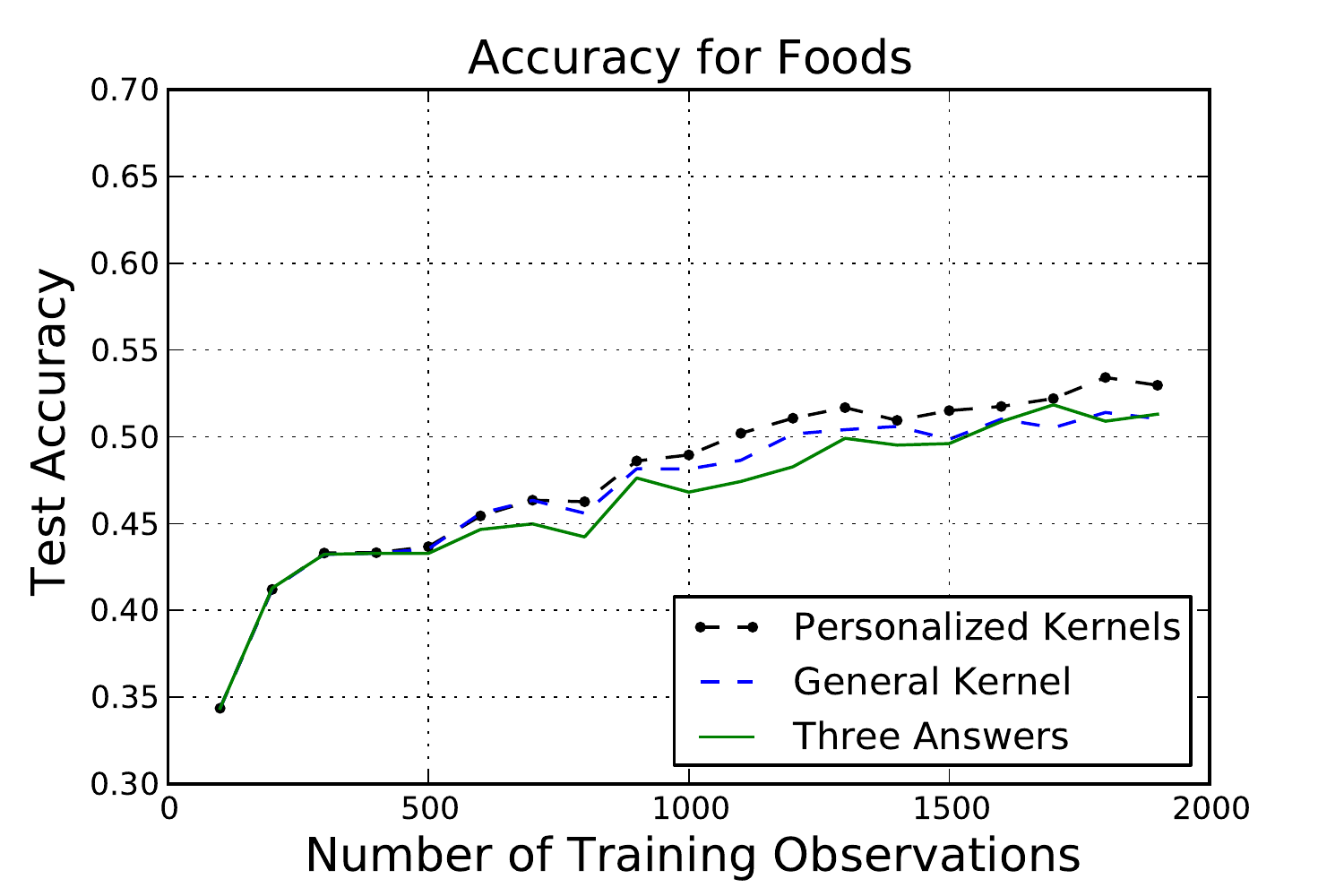}
\quad
\includegraphics[width=0.5\textwidth]{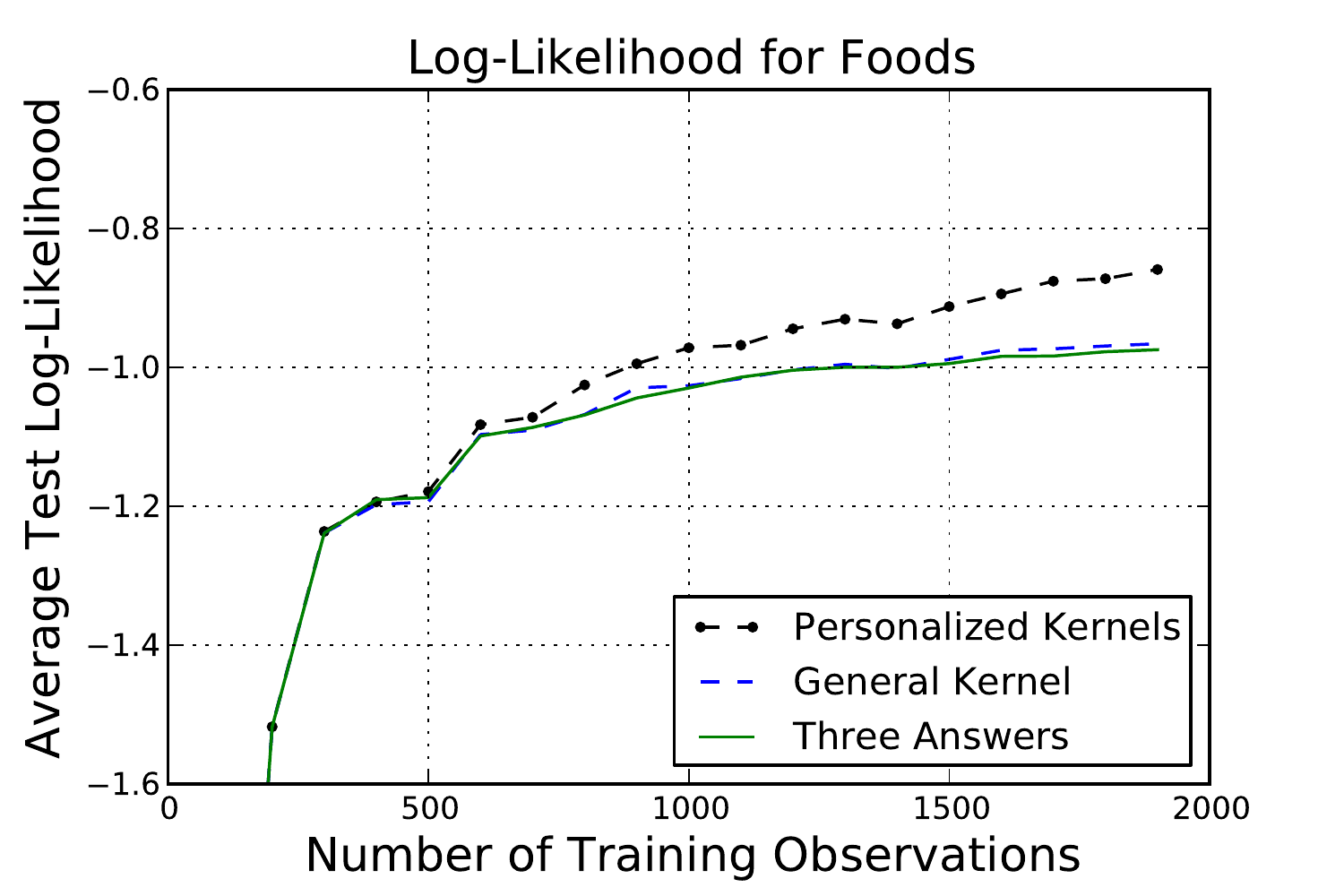}
\caption{Learning curves for user models on foods data. The personalized kernel has a higher log-likelihood for the
test data, but as in our prior plot this does not translate to a much higher test accuracy.}
\label{user-accuracy-foods}
\end{figure}

\section{Future Work}

We believe this framework has a great deal of potential, particularly in the domain of constructing
evaluation collections for machine learning and information retrieval tasks.  We see several opportunities
for its further development.
\begin{itemize}
\item Use on text collections: the current technique is best suited to notions of similarity that can be easily
captured by simple questions that people can answer without training. This makes it somewhat ill-suited to use
on several useful notions of text similarity, such as entailment or query relevance. We seek to develop the
framework further to target these.
\item Active learning: We have not modified the active learning process from the original, but we see an
opportunity for improvement here. The process considers only a few discrete future object positions given some
question it's considering, and we find empirically that it often chooses sub-optimal questions.
\item Scalability: Our current model is often still learning after asking 20 questions per head. This is many
fewer questions than the total $n \cdot {n-1 \choose 2}$, but still difficult to scale to very
large object collections. We would like to address this, perhaps by quickly finding a large-scale positioning of all
objects and then considering only questions which position objects locally. Another possible approach is to find ways
to incorporate existing data collected from slightly different questions, such as pair preferences or clickthrough data.
\item Better optimization: the optimization method we're using is somewhat prone to getting trapped in local
minima. We believe that this can be avoided by either using a better optimization technique, or perhaps by
adjusting the model to be somewhat more convex.
\end{itemize}

\section{Conclusion}

We have presented an extension of the Crowd Kernel model by Tamuz et al which corrects
for two major sources of noise: users' inability to compare dissimilar objects, and
inter-assessor disagreement between users. This allows this useful model to be applied
more effectively on heterogeneous data sets.
Our results showed the most dramatic improvements for data that falls more or less into distinct
classes, with relatively few objects defying categorization. However, the model improves somewhat over the
original even for our more ambiguous data set.

\bibliographystyle{abbrv}
\bibliography{aslam,ir,iis10,local}

\newpage
\appendix
\label{sec:appendix}
We perform Newton-Raphson approach to optimize for the object embeddings ${M_x}$ and parameters $\Omega$:
\begin{align*}
\{M, \Omega\} := \{M, \Omega\} - H^{-1}\nabla_{\{M, \Omega\}} \sum_{i = 1}^{n} \ln \hat{p}(r_i | M_{a_i}, M_{b_i}, M_{c_i}, \Omega)
\end{align*}
To speed up computation, we consider Hessian matrix with zeroes outside the main diagonal.

Assuming 
\begin{align*}
\lambda&=1 \\
c_1&=(\mu_k + \delta_{ab}^k)^{-1} \\
c_2&=(\mu_k + \delta_{ac}^k)^{-1} \\
c_3&=(\mu_k + d^2_k + \delta_{ab}^k)^{-1} \\
c_4&=(\mu_k + d^2_k + \delta_{ac}^k)^{-1} \\
c_5&=(\mu_k + \delta_{ac}^k)^{-1} \\
c_6&=(2 + \delta_{ab}^k + \delta_{ac}^k)^{-1} \\
c_7&=(d^2_k)^{-1} \\
c_8&=(2\mu_k + d^2_k + \delta_{ab}^k + \delta_{ac}^k)^{-1} 
\end{align*}

\begin{align*}
\frac{\partial \ln \hat{p}^a_{bc}}{\partial \delta_{ab}^k}&=c_8 - c_3 - c_6 \\
\frac{\partial \ln \hat{p}^a_{bc}}{\partial \delta_{ac}^k}&=c_8 - c_4 + c_5 - c_6 \\
\frac{\partial \ln \hat{p}^a_{bc}}{\partial d^2_k}&=c_7 + c_8 - c_3 - c_4 \\
\frac{\partial \ln \hat{p}^a_{bc}}{\partial \mu_k}&=2c_8 - c_3 - c_4 
\end{align*}

\begin{align*}
\frac{\partial \ln \hat{p}_{neither}}{\partial \delta_{ab}^k}&=c_1 - c_3 \\
\frac{\partial \ln \hat{p}_{neither}}{\partial \delta_{ac}^k}&=c_2 - c_4 \\
\frac{\partial \ln \hat{p}_{neither}}{\partial d^2_k}&= -c_3 - c_4 \\
\frac{\partial \ln \hat{p}_{neither}}{\partial \mu_k}&=c_1 + c_2 - c_3 - c_4 
\end{align*}

\begin{align*}
\frac{\partial^2 \ln \hat{p}^a_{bc}}{(\partial \delta_{ab}^k)^2}&=-(c_8)^2 + (c_3)^2 + (c_6)^2 \\
\frac{\partial^2 \ln \hat{p}^a_{bc}}{(\partial \delta_{ac}^k)^2}&=-(c_8)^2 + (c_4)^2 - (c_5)^2 + (c_6)^2 \\
\frac{\partial^2 \ln \hat{p}^a_{bc}}{(\partial d^2_k)^2}&=-(c_7)^2 - (c_8)^2 + (c_3)^2 + (c_4)^2 \\
\frac{\partial^2 \ln \hat{p}^a_{bc}}{(\partial \mu_k)^2}&=-4(c_8)^2 + (c_3)^2 + (c_4)^2 
\end{align*}

\begin{align*}
\frac{\partial^2 \ln \hat{p}_{neither}}{(\partial \delta_{ab}^k)^2}&=-(c_1)^2 + (c_3)^2 \\
\frac{\partial^2 \ln \hat{p}_{neither}}{(\partial \delta_{ac}^k)^2}&=-(c_2)^2 + (c_4)^2 \\
\frac{\partial^2 \ln \hat{p}_{neither}}{(\partial d^2_k)^2}&=(c_3)^2 + (c_4)^2 \\
\frac{\partial^2 \ln \hat{p}_{neither}}{(\partial \mu_k)^2}&=-(c_1)^2 - (c_2)^2 + (c_3)^2 + (c_4)^2 
\end{align*}

\begin{align*}
\frac{\partial \delta_{xy}^k}{\partial M_x^d}&=2u_k^d(M_x^d - M_y^d) \\
\frac{\partial \delta_{xy}^k}{\partial M_y^d}&=-2u_k^d(M_x^d - M_y^d) \\
\frac{\partial \delta_{xy}^k}{\partial u_k^d}&=(M_x^d - M_y^d)^2 
\end{align*}

\begin{align*}
\frac{\partial^2 \delta_{xy}^k}{(\partial M_x^d)^2}&=2u_k^d \\
\frac{\partial^2 \delta_{xy}^k}{(\partial M_y^d)^2}&=2u_k^d \\
\frac{\partial^2 \delta_{xy}^k}{(\partial u_k^d)^2}&=0 
\end{align*}

\begin{align*}
\frac{\partial \ln \hat{p}}{\partial M_a^d}&=\frac{\partial \ln \hat{p}}{\partial \delta_{ab}^k}*\frac{\partial \delta_{ab}^k}{\partial M_a^d} + \frac{\partial \ln \hat{p}}{\partial \delta_{ac}^k}*\frac{\partial \delta_{ac}^k}{\partial M_a^d} \\
\frac{\partial \ln \hat{p}}{\partial M_b^d}&=\frac{\partial \ln \hat{p}}{\partial \delta_{ab}^k}*\frac{\partial \delta_{ab}^k}{\partial M_b^d} \\
\frac{\partial \ln \hat{p}}{\partial M_c^d}&=\frac{\partial \ln \hat{p}}{\partial \delta_{ac}^k}*\frac{\partial \delta_{ac}^k}{\partial M_c^d} \\
\frac{\partial \ln \hat{p}}{\partial u_k^d}&=\frac{\partial \ln \hat{p}}{\partial \delta_{ab}^k}*\frac{\partial \delta_{ab}^k}{\partial u_k^d} + \frac{\partial \ln \hat{p}}{\partial \delta_{ac}^k}*\frac{\partial \delta_{ac}^k}{\partial u_k^d} 
\end{align*}

\begin{align*}
\frac{\partial^2 \ln \hat{p}}{(\partial M_a^d)^2}&=\frac{\partial^2 \ln \hat{p}}{(\partial \delta_{ab}^k)^2}*\left(\frac{\partial \delta_{ab}^k}{\partial M_a^d}\right)^2 + \frac{\partial \ln \hat{p}}{\partial \delta_{ab}^k}*\frac{\partial^2 \delta_{ab}^k}{(\partial M_a^d)^2} \\
&+ \frac{\partial^2 \ln \hat{p}}{(\partial \delta_{ac}^k)^2}*\left(\frac{\partial \delta_{ac}^k}{\partial M_a^d}\right)^2 + \frac{\partial \ln \hat{p}}{\partial \delta_{ac}^k}*\frac{\partial^2 \delta_{ac}^k}{(\partial M_a^d)^2} \\
\frac{\partial^2 \ln \hat{p}}{(\partial M_b^d)^2}&=\frac{\partial^2 \ln \hat{p}}{(\partial \delta_{ab}^k)^2}*\left(\frac{\partial \delta_{ab}^k}{\partial M_b^d}\right)^2 + \frac{\partial \ln \hat{p}}{\partial \delta_{ab}^k}*\frac{\partial^2 \delta_{ab}^k}{(\partial M_b^d)^2} \\
\frac{\partial^2 \ln \hat{p}}{(\partial M_c^d)^2}&=\frac{\partial^2 \ln \hat{p}}{(\partial \delta_{ac}^k)^2}*\left(\frac{\partial \delta_{ac}^k}{\partial M_c^d}\right)^2 + \frac{\partial \ln \hat{p}}{\partial \delta_{ac}^k}*\frac{\partial^2 \delta_{ac}^k}{(\partial M_c^d)^2} \\
\frac{\partial^2 \ln \hat{p}}{(\partial u_k^d)^2}&=\frac{\partial^2 \ln \hat{p}}{(\partial \delta_{ab}^k)^2}*\left(\frac{\partial \delta_{ab}^k}{\partial u_k^d}\right)^2 + \frac{\partial \ln \hat{p}}{\partial \delta_{ab}^k}*\frac{\partial^2 \delta_{ab}^k}{(\partial u_k^d)^2} \\
&+ \frac{\partial^2 \ln \hat{p}}{(\partial \delta_{ac}^k)^2}*\left(\frac{\partial \delta_{ac}^k}{\partial u_k^d}\right)^2 + \frac{\partial \ln \hat{p}}{\partial \delta_{ac}^k}*\frac{\partial^2 \delta_{ac}^k}{(\partial u_k^d)^2} 
\end{align*}

\end{document}